\documentclass{article}
\usepackage{spconf,amsmath,graphicx}

\title{Optimal Sample Lens Positioning in Digital Camera Systems}
\name{Ali Karaoglu 
}
\address{akaraoglu@gmail.com}
\begin{document}
%
\maketitle
\begin{abstract}
In contemporary imaging systems, achieving optimal autofocus (AF) performance hinges on precise lens positioning. Extensive research has delved into refining algorithms for determining the ideal lens position across passive, active, and hybrid autofocus systems. This paper explores the mathematical intricacies and practical considerations essential for optimizing lens positions during focus searches, addressing overarching challenges encountered in AF systems, such as balancing speed and accuracy. Moreover, the lens position calculations proposed herein can be applied to various focus algorithms, including focus bracketing. The proposed method offers adaptability and scalability, rendering it suitable for integration into a wide array of camera systems, ranging from smartphones and DSLRs to microscopes and industrial imaging devices.
\end{abstract}
\section{Introduction}
\label{sec:intro}

The primary goal of focusing is to ensure that the subject of the image is rendered with maximum sharpness. Whether manually or through autofocus systems, achieving correct focus is essential for capturing images that are visually appealing and effectively communicate the photographer's intent. In scenarios ranging from macro photography to landscape and portrait photography, proper focus determines whether the details necessary for the image's impact are adequately captured.

Autofocus (AF) systems are designed to automatically adjust the camera lens to achieve clear, sharp images of subjects by focusing the lens correctly without manual intervention. These systems are essential components in many modern cameras, including DSLRs, mirrorless cameras, and smartphones, providing convenience and speed that are critical in various photography scenarios. AF systems operate by receiving data from the  sensor, processing this data through an autofocus algorithm, and then directing motor systems in the camera module to adjust the lens elements until the subject is in focus. These systems can operate in single-shot modes, where focus is achieved once when the shutter button is partially pressed, or in continuous modes \cite{Gamadia2006}, which constantly adjust focus.

The accuracy of autofocus systems hinges significantly on the determination of optimal lens positions during the focusing process. Sample lens positions, when judiciously chosen, can dramatically reduce the focusing time and increase the accuracy by limiting the search space and focusing efforts on the most likely regions for achieving sharp focus. This is particularly critical in environments where the subject's distance varies unpredictably.

In this paper, we propose a method to calculate the necessary sample lens positions using given set of parameters. This method can be used variety of focusing operations such as hill climbing methods, fine tuning the focus position for hybrid solutions or focus bracketing algorithm.

\section{Related Work}
\label{sec:relatedwork}

Numerous autofocus algorithms have been developed and studied to optimize the process of achieving clear and sharp images. Techniques such as phase detection \cite{Chan:ICIP17}, contrast detection \cite{he2003modified}, and hybrid autofocus algorithms \cite{herrmann2020learning} have been investigated to improve the speed and accuracy of focusing mechanisms. Moreover, focus bracketing algorithms \cite{araujo2022focus} have been studied to increase the depth of field by focus stacking. 

Previous studies focused on improving performance and overcoming obstacles primarily within two central components of traditional contrast maximization methods: the evaluation metric and the search strategy. \cite{Yao2006Evaluation} \cite{gamadia2010performance}. 

Evaluation metrics, which are specially designed features, assess the sharpness of an image to gauge its quality. When an evaluation metric is established, the goal of a search strategy is to optimize this metric by precisely identifying the best focus position. Extensive research has been carried out on both these aspects in scholarly studies. There are a number of non-learning techniques in the image processing literature,
\cite{Yao2006Evaluation}, \cite{kehtarnavaz2003development}, \cite{he2003modified}, 
\cite{guo2018fast}, \cite{gamadia2012filter}, \cite{yousefi2011new}, 
\cite{gamadia2010performance}.\cite{Lee2008}, \cite{Lee2009}, \cite{Yang:ICIP16}, \cite{Yang:TIP18}], 
and learning based techniques  
\cite{wang2020intelligent}, \cite{park2008fast}, \cite{chen2010passive}, \cite{han2011novel}, \cite{mir2015autofocus}, \cite{jiang2018transform}. 
CNN based algorithms \cite{Nanda01} \cite{herrmann2020learning} are also studied to estimate the optimal focus position using the information obtained from single or multiple images. 

The shortcomings of traditional contrast maximization methods are:
\begin{itemize}
    \item The evaluation metric does not directly indicate the focus state of the image.
    \item The movement distance at each time step is predetermined by the search strategy.
\end{itemize}

These limitations suggest that an intelligent autofocus (AF) system should be capable of assessing whether an image is in focus and, if it is not, determining the necessary travel distance to achieve optimal focus.

Traditional systems employ a search strategy to optimize the autofocus (AF) metric, ideally minimizing the number of lens movement(steps) required in the process. Popular search strategies include Fibonacci \cite{krotkov1988}, rule-based \cite{kehtarnavaz2003development} and hill-climbing \cite{he2003modified}. To decrease the number of steps, more complex algorithms such as curve fitting ROL \cite{yazdanfar2008simple} and BPIC \cite{wu2012bilateral} were proposed. 

Optimizing and evaluating the AF process by using traditional AF, ML-based AF and Focus Bracketing/Stacking \cite{wang2020intelligent} for a variety of scenarios and use cases have been widely covered in the image processing literature. 

However, every camera module designed with different focal length, aperture and circle of confusion has different depth of field and sharpness accuracy in the final image. The traditional and ML based studies show that optimizing the number lens steps are hand tuned and evaluated by some sharpness measurement. Since each camera module is different, this process needs to be done for every camera and will include many rounds of image quality assessments to tune the algorithms. Additionally, to achieve the same focus quality in different image resolution or aperture size is not taken into account. 

\section{Proposed Method}
\label{sec:proposed}

In this paper we propose to calculate the lens steps with respect to the module and optimize the number of steps according to the final quality requirements. 
To understand the relation between camera module and AF systems, first we need go through basics. 

\begin{figure}
\centering
\includegraphics[width=0.6\linewidth]{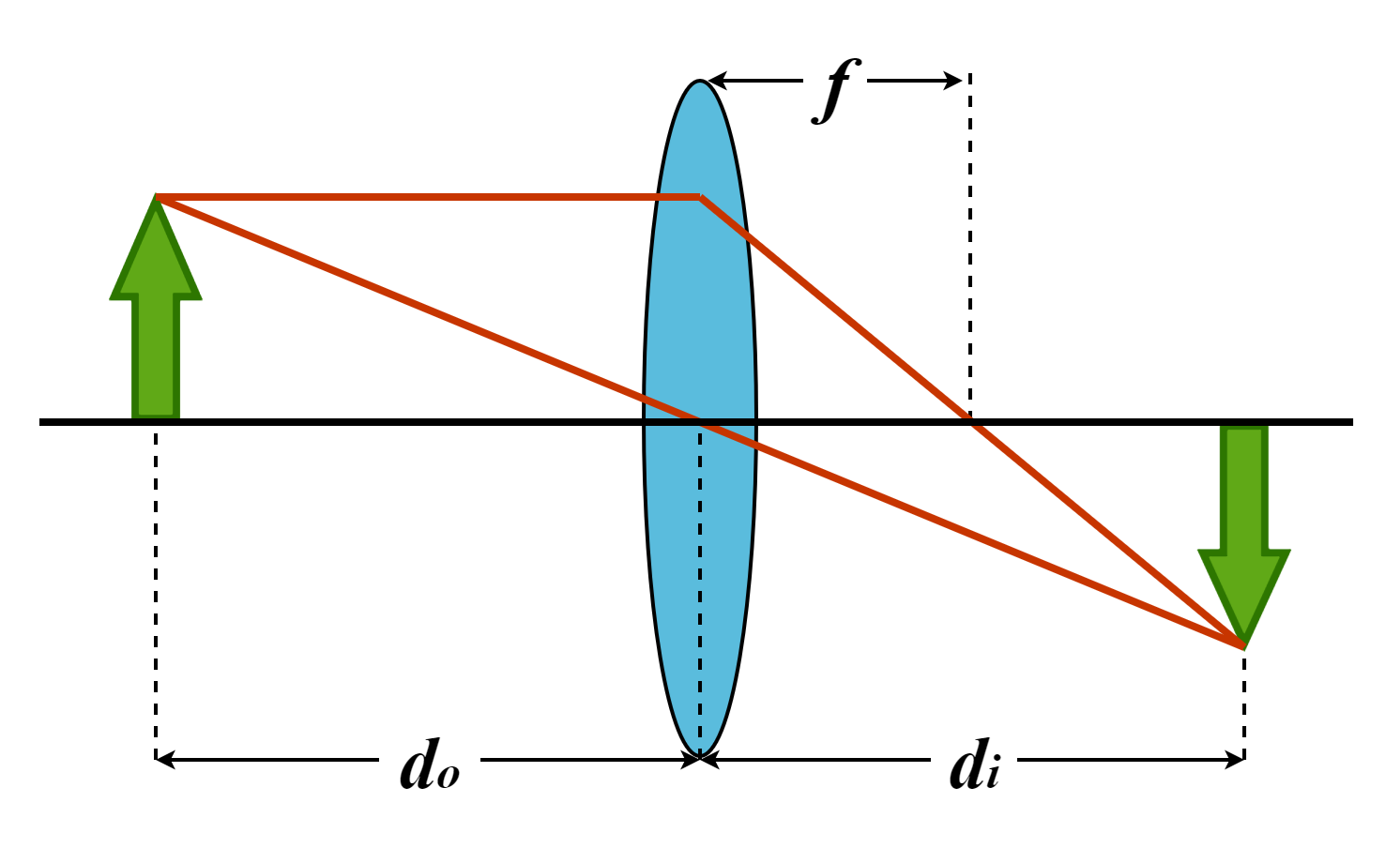}
\caption{\label{fig:thinlens}Thin lens example, shows the relationship between the object distance and the image distance.}
\end{figure}

\subsection{Basic Principles of Optical Systems}
\subsubsection{Thin-Lens Equation}
\label{sec:thinlens}

The Thin-Lens Equation is a fundamental principle in optics that describes the relationship between the focal length of a lens, the distance from the lens to the object being focused on, and the distance from the lens to the image formed by that object. See Figure \ref{fig:thinlens}. It applies to lenses that are thin enough that their thickness can be neglected in calculations, a common approximation in many practical situations.

The formula for the Thin-Lens Equation is:
\[
\frac{1}{f} = \frac{1}{d_o} + \frac{1}{d_i}
\]
Where  \( f \)  is the focal length, \( d_o \) is the distance to the object, and \( d_i \) is the distance to the image plane (sensor). This fundamental formula helps determine where the lens should be focused for a clear image of an object at a given distance.

\begin{figure}
\centering
\includegraphics[width=0.8\linewidth]{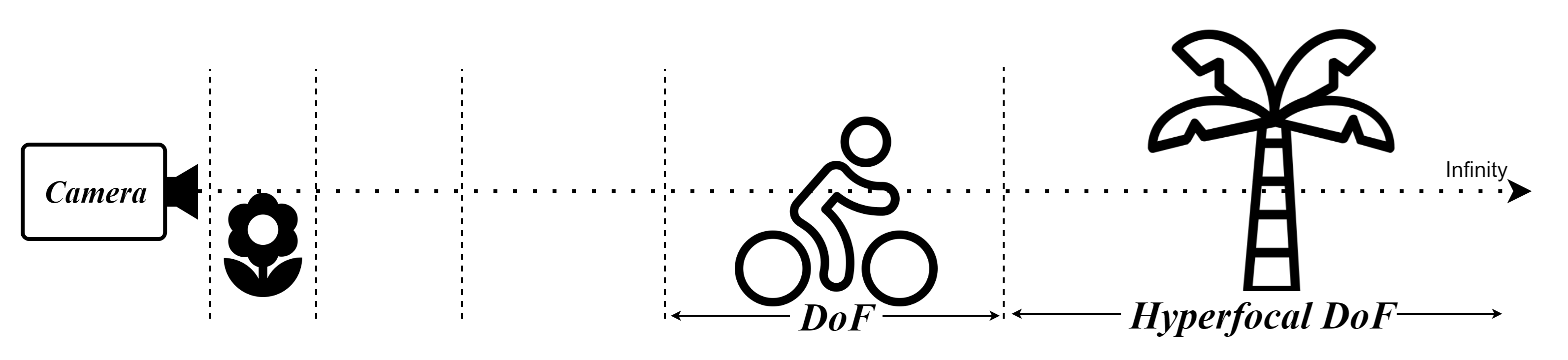}
\caption{\label{fig:dof} A scene can be split into many depth of fields. Note that the number of DoFs and relation between them in the figure is fictional. The hyperfocal DoF is the DoF when the lens is focused to  hyperfocal distance.}
\end{figure}

\subsubsection{Circle of Confusion}
\label{sec:coc}

\begin{figure}
\centering
\includegraphics[width=0.8\linewidth]{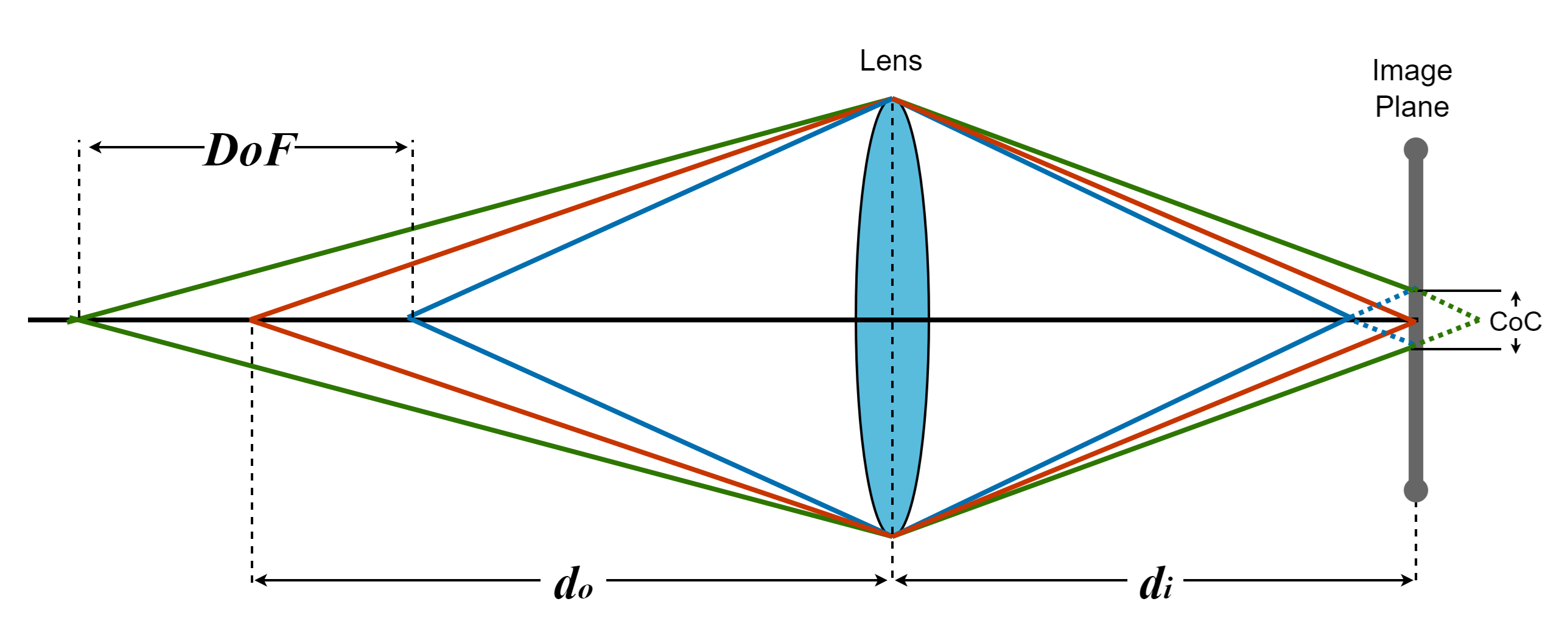}
\caption{\label{fig:coc} The depth of field in the scene can be visualised over the image plane as shown. The circle of confusion decides the size of depth of field that is allowed to be in sharp in the image. Blue line shows the near end of the DoF, green line show the far end of the DoF. }
\end{figure}

The Circle of Confusion (CoC) refers to the optical spot created by a cone of light from a lens that is not perfectly focused. See Figure \ref{fig:coc} In simpler terms, it is the degree to which a point light source appears as a blurred, disc-like shape in an image, which affects image sharpness. CoC is a critical concept in understanding optical imaging, particularly in photography and cinematography. It represents the maximum blur spot that is indistinguishable from a point to the human eye or is considered acceptably sharp in an image. CoC is influenced by sensor size, viewing conditions, and print size.

The Circle of Confusion can be calculated based on sensor size and desired print resolution. A commonly used formula to estimate CoC is:
\[
CoC = \frac{d}{k}
\]
where \( d \) is the diagonal of the sensor in millimeters \(k\) and is the constant that comes from typical viewing conditions and visual acuity assumptions (Print Size / Image Size).
This constant can vary depending on the source, with some using 1500 to 2000 based on more critical sharpness criteria. In the industry it is usually set as 1730.  Note that by changing \(k\), one can adjust the CoC, which will impact the final quality of the sharpness of focusing.

\subsubsection{Hyperfocal Distance}
\label{ssec:hyperfocal}

Hyperfocal distance is the closest distance at which a lens can be focused while keeping objects at infinity acceptably sharp; focusing at this distance maximizes the depth of field from half this distance to infinity. See Figure \ref{fig:dof}. 
The Hyperfocal Distance \( H \) can be calculated using the formula:
\[
H = \frac{f^2}{N \cdot c}
\]
where \( f \) is the focal length of the lens, \( N \) is the aperture number (f-stop), and \( c \) is the Circle of Confusion.
By focusing on the hyperfocal distance, photographers can achieve the deepest possible depth of field for a given aperture and focal length. The near and far focus limits are particularly useful for landscape photographers who need to ensure maximum sharpness from the foreground to the horizon. Understanding and calculating these distances allows photographers to pre-plan their shots for optimal sharpness across the entire image.

\subsubsection{Depth of Field}
\label{sssec:dof}
Depth of Field (DoF) is the range within a photo that appears acceptably sharp. See Figure \ref{fig:dof}.  It is influenced by the lens aperture, the focal length, the focus distance, and the size of the Circle of Confusion. It is the range from near focus distance to far focus distance. The formula for calculating DoF is given by:
The Depth of Field (DoF) is given by the formula:
\[
DoF = \frac{2 d_o N c f^2}{f^4 - (N c d_o)^2}
\]
where:
\begin{itemize}
    \item \( d_o \) is the distance to the object in focus,
    \item \( N \) is the aperture number (f-stop),
    \item \( c \) is the Circle of Confusion,
    \item \( f \) is the focal length.
\end{itemize}

\subsubsection{Limits of acceptable sharpness}
\label{sec:limits}
The near focus and far focus distance limits define the boundaries within which all objects are rendered acceptably sharp in an image.

The Near Focus Distance Limit \( D_n \) is calculated using the formula:
\[
D_n = \frac{d_o \cdot (H-f)}{H + d_o -2f}
\]
where:
\begin{itemize}
    \item \( d_o \) is the distance to the object in focus,
    \item \( H \) is the hyperfocal distance,
    \item \( f \) is the focal length.
\end{itemize}

The Far Focus Distance Limit \( D_f \) is given by the formula:
\[
D_f = \frac{d_o \cdot (H-f)}{H - d_o}
\]
If \( d_o \) is greater than \( H \), then \( D_f \) is infinity.

\subsection{Method}
\label{sec:method}

The near and far object distances tell us that the range of DoF in focus according to the physical properties of the module. With this information we can slice the focus range into separate depth of fields, which will help us to cover a focus range with given quality parameters.  Slicing the focus range can be done to both directions, towards near or far.  Our calculations will be using the direction from near end towards infinity. Therefore, we need to start the slicing operations from a shortest distance to the lens were the camera can focus. This distance is usually defined by the autofocus lens movement and has a practical limit. It is recommended to choose an initial \( d_o \) distance a bit further than the \( f \) number of the lens. 

Using \( d_o \), the far distance of the depth of field can be calculated using formula shown in \ref{sssec:dof} and far limit location will be used as the near limit of the next DoF slice. Therefore, \( D_f \) is the next \( D_n \).

To calculate the \( D_f \) from given \( D_n \), first the new \( d_o \) is calculated.
We derive the formula of finding the focus distance using near focus limit: 
\[
d_o =  \frac{(D_n * (H - 2 * F))}{(H - F - D_n)}
\]

Using the formula above, we obtain the next \( d_o \) and that is the next sample focus distance. The process will continue as a loop until, first calculate \( D_f \) and then calculate the next \( d_o \). At this point, one can realize that the range may not be sliced into DoFs evenly. Thus, we need to decide place of the overlap in the focus range. For our proposes, this will in the furthest lens position. Thus we stop the loop when the  \( D_f \) is in the furthest lens position's DoF, which can be hyperfocal or a specified distance. 

To calculate the same operation starting from far focus limit toward near end use the formula:
\[
d_o =  \frac{(D_f * H)}{(H - F + D_f)}
\]

The sample focus distances are now calculated per each  \( DoF \) and covering the whole range with no gap. The next step is to convert the distances to image plane. As shown in Figure \ref{fig:thinlens}, the distance from the object to the lens in the real world is related to the image distance from the lens. From a given distance of the focused object in the real world, we can calculate the lens position and vice versa using the thin lens formula shown in \ref{sec:thinlens}.
\[
d_i = \frac{(d_o*f)} {(d_o-f)}
\]
\[
d_o = \frac{(d_i*f)} {(d_i-f)}
\]
where:
\begin{itemize}
    \item \( d_o \) is the distance to the object in focus,
    \item \( d_i \) is the distance to the image in focus,
\end{itemize}

After converting the distances, we determine the image plane distances essential for the focusing algorithm. The next stage, not discussed in this paper, involves controlling the camera module's lens. This process varies with the lens and device specifications, and developers are expected to have a thorough understanding of the lens's operational capabilities and limitations.

These operations that are shown above are optimized for the  \( CoC \) that is given as an input to the equation in hyperfocal  \( H \)  distance calculation. The  \( CoC \) is directly proportional to the  \( DoF \) range and it will make the algorithm calculate longer distances. It is the parameter that controls the desired quality in the final image. The same parameter can be tuned for focus bracketing algorithms to calculate the lens positions that are required for the stacking algorithm. Tuning the algorithm can be done easily with deciding an optimal  \( CoC \) for the algorithm and our proposed method can find the optimal lens position. 

The operations detailed above are tailored to optimize the Circle of Confusion (CoC), which serves as an input for calculating the hyperfocal distance. The CoC is directly proportional to the Depth of Field (DoF) range, influencing the algorithm to calculate longer distances. It serves as the key parameter in determining the desired image quality. This parameter can also be adjusted for focus bracketing algorithms to identify the necessary lens positions for the stacking algorithm. Our proposed method simplifies this tuning by selecting an optimal CoC, enabling precise determination of the best lens position.

\subsection{Experiment}
\label{sec:experiment}
A simple experiment with the proposed method will easily provide the sample lens positions, following the operations and results.

First we assume the initial parameters as: 

\begin{itemize}
    \item Focal Length (\( F\)): 25.000 mm,
    \item Aperture (\( N\)): f/4.6
    \item Circle of Confusion (\( c\)): 0.0200 mm
    \item Practical Nearest Focus Distance (\( S_n\) ): 250 mm
\end{itemize}

Using these parameters and the formulas for Hyperfocal distance \ref{ssec:hyperfocal} and thin-lens \ref{sec:thinlens} equations, we obtain: 

\begin{itemize}
    \item Hyperfocal Distance (\( H\)): 6818.48 mm
    \item Lens Distance at Nearest Focus Limit: 27.78 mm
    \item Lens Distance at Hyperfocal Distance: 25.09 mm
\end{itemize}

After the initial calculations, the sample lens positions can be obtained in a loop shown in  Table \ref{tab:samplelenspos}.

\begin{table}
    \centering
    \begin{tabular}{ccccc}
        Step & Focus Distance in mm & Lens Distance in mm\\
        0&267.8&27.5741\\
        1&288.5&27.3719\\
        2&312.9&27.1712\\
        3&342.0&26.9719\\
        4&377.3&26.7741\\
        5&421.1&26.5778\\
        6&476.9&26.3829\\
        7&550.5&26.1894\\
        8&651.6&25.9974\\
        9&799.7&25.8067\\
        10&1037.1&25.6175\\
        11&1479.7&25.4297\\
        12&2595.1&25.2432\\
        13&6817.5&25.0920\\
    \end{tabular}
    \caption{Calculated sample lens positions for the full focus range to capture each focus distance in focus.}
    \label{tab:samplelenspos}
\end{table}

\subsection{Conclusion}
\label{sec:conclusion}

Understanding the physical properties of a camera module significantly impacts focusing algorithms. The proposed method in this paper comprehensively considers these properties, allowing for the straightforward calculation of sample lens positions using the provided formulas. The benefits of this algorithm begin with its use of the final image resolution, which directly influences the optimal lens position. For lower resolutions, a coarser lens position suffices for accurate focus.

Furthermore, the speed and accuracy of determining focus distances are crucial in a contrast-based hill-climbing algorithm. The calculations detailed in this document simplify the tuning and understanding of a focusing algorithm, enhancing both elements. Additionally, the accuracy of active autofocus (AF) systems is measured and calibrated. Given their limitations with high-resolution images, we have incorporated a passive fine-search autofocus algorithm to improve precision.

Another advancement is our ability to calculate lens positions for focus bracketing from a set of parameters within a specified range. By mastering these concepts and applying the correct formulas, photographers can capture optimal amount of images with the best depth and sharpness. Our method significantly simplifies to find the best focus bracketing sample lens locations for the best photographic quality.

Looking ahead, a key area for improvement in our focusing algorithm involves incorporating scene content analysis. Currently, our algorithm does not account for the specifics of the scene, which could significantly refine the optimization of focus steps. By integrating scene analysis, the algorithm could dynamically adjust focus based on the elements present in the scene, such as distinguishing between foreground and background or recognizing specific subjects like faces or text. This enhancement would allow for more precise and context-sensitive focusing, potentially improving both the speed and accuracy of the autofocus system in varied photographic conditions.




\bibliographystyle{IEEEbib}
\bibliography{strings,refs}

\end{document}